\RequirePackage{fix-cm}

\documentclass[twocolumn]{svjour3}

\smartqed  

\usepackage{graphicx}%

\usepackage{amsmath}
\usepackage{algorithm}
\usepackage{algpseudocode}  
\usepackage{amsfonts}
\usepackage{enumerate}
\usepackage[accsupp]{axessibility}

\usepackage{natbib}

\usepackage{ragged2e}
\usepackage{appendix}
\usepackage{booktabs}
\usepackage{color}
\usepackage{multirow}
\usepackage{bm}
\usepackage{bbm}
\usepackage{caption}
\usepackage[labelformat=simple]{subcaption}
\usepackage{hyperref}
\hypersetup{
    colorlinks=true,
    linkcolor=red,
    filecolor=blue,      
    urlcolor=red,
    citecolor=blue,
}

\usepackage{colortbl}
\usepackage{booktabs}

\definecolor{Gray1}{rgb}{0.92,0.92,0.92}
\definecolor{Gray2}{rgb}{0.88,0.88,0.88}
\newcommand{\parsection}[1]{\noindent\textbf{#1:}~}

\begin{document}\sloppy

\title{I$^2$MD: 3D Action Representation Learning with Inter- and Intra-modal Mutual Distillation}

\author{Yunyao Mao\and
        Jiajun Deng\and
        Wengang Zhou\and
        Zhenbo Lu\and
        Wanli Ouyang\and
        Houqiang Li
}

\authorrunning{Yunyao Mao,~\emph{et al.}} 

\institute{
Yunyao Mao \at
CAS Key Laboratory of Technology in GIPAS, Department of Electronic Engineeringand Information Science, University of Science and Technology of China, Hefei, China. \\
\email{myy2016@mail.ustc.edu.cn}
\and
Jiajun Deng \at
The University of Adelaide, Adelaide, Australia. \\
\email{jiajun.deng@adelaide.edu.au}
\and
Wengang Zhou \at
CAS Key Laboratory of Technology in GIPAS, Department of Electronic Engineeringand Information Science, University of Science and Technology of China, Hefei, China. \\
\email{zhwg@ustc.edu.cn}
\and
Zhenbo Lu \at
Institute of Artificial Intelligence, Hefei Comprehensive National Science Center, Hefei, China. \\
\email{luzhenbo@iai.ustc.edu.cn}
\and
Wanli Ouyang \at
Shanghai Artificial Intelligent Laboratory, Shanghai, China. \\
\email{wanli.ouyang@sydney.edu.au}
\and
Houqiang Li \at
CAS Key Laboratory of Technology in GIPAS, Department of Electronic Engineeringand Information Science, University of Science and Technology of China, Hefei, China. \\
\email{lihq@ustc.edu.cn}
\and
Corresponding Authors: Jiajun Deng and Wengang Zhou
}

\date{Received: date / Accepted: date}

\maketitle
\begin{abstract}
Recent progresses on self-supervised 3D human action representation learning are largely attributed to contrastive learning. However, in conventional contrastive frameworks, the rich complementarity between different skeleton modalities remains under-explored. Moreover, optimized with distinguishing self-augmented samples, models struggle with numerous similar positive instances in the case of limited action categories. In this work, we tackle the aforementioned problems by introducing a general Inter- and Intra-modal Mutual Distillation (I$^2$MD) framework. In I$^2$MD, we first re-formulate the cross-modal interaction as a Cross-modal Mutual Distillation (CMD) process. Different from existing distillation solutions that transfer the knowledge of a pre-trained and fixed teacher to the student, in CMD, the knowledge is continuously updated and bidirectionally distilled between modalities during pre-training. To alleviate the interference of similar samples and exploit their underlying contexts, we further design the Intra-modal Mutual Distillation (IMD) strategy, In IMD, the Dynamic Neighbors Aggregation (DNA) mechanism is first introduced, where an additional cluster-level discrimination branch is instantiated in each modality. It adaptively aggregates highly-correlated neighboring features, forming local cluster-level contrasting. Mutual distillation is then performed between the two branches for cross-level knowledge exchange. Extensive experiments on three datasets show that our approach sets a series of new records.

\end{abstract}



\section{Introduction}
\label{sec:introduction}

As one of the fundamental problems in computer vision, human action recognition has been widely adopted in varieties of applications, such as behavior analysis, human-machine interaction, virtual reality, \emph{etc}. 
Nowadays, with the advancement of pose estimation techniques \citep{openpose,fang2017rmpe,xu2020deep,zhang2023learning}, skeleton-based 3D human action recognition \citep{gupta2021quo} has attracted increasing research interests thanks to its lightweight and background-robust characteristics.  
Nevertheless, fully-supervised 3D action recognition \citep{8936339,9076822,Chen_2021_ICCV,du2015hierarchical,ke2017new,li2019actional,Li_2021_ICCV,liu2020disentangling,shi2019skeleton,Shi_2021_ICCV,si2019attention,zhang2019view,zhang2020semantics,zhang2020context} requires large amounts of well-annotated skeleton data, which is rather labor-intensive to acquire. 
Such facts motivate and highlight the explorations of self-supervised 3D action representation learning. 
In literature, the early works~\citep{lin_ms2l_mm,misra2016shuffle,nie2020unsupervised,noroozi2016unsupervised,su2020predict,zheng2018unsupervised,nie2021view,9623511} devise various celebrated pretext tasks like motion prediction, jigsaw puzzle recognition, and masked reconstruction.
Recently, the contrastive learning paradigm \citep{chen2020simple,he2020momentum,van2018representation} has been introduced to skeleton-based 3D action recognition \citep{lin_ms2l_mm,rao2021augmented,9612062}. It achieves great success thanks to the capability of learning discriminative high-level semantic features.

\begin{figure*}[t!]
    \centering
    \includegraphics[width=1.0\linewidth]{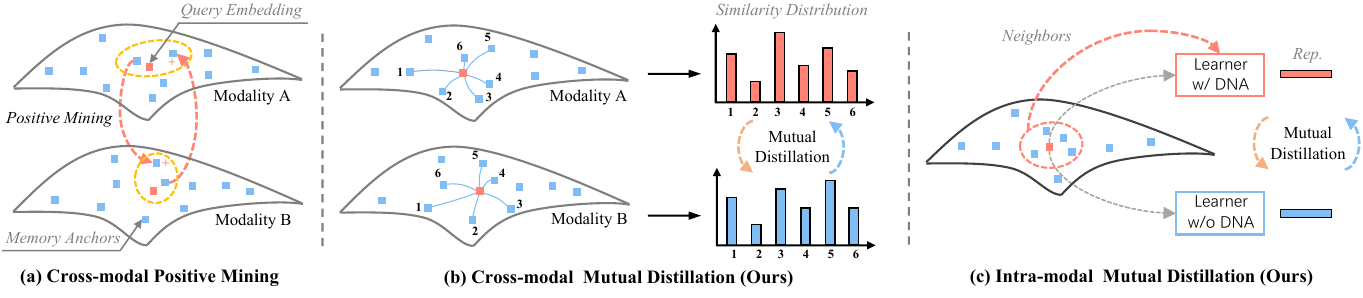}
    \caption{An illustration of Cross-modal Positive Mining (CPM) \citep{li20213d}, Cross-modal Mutual Distillation (CMD), and Intra-modal Mutual Distillation (IMD). Given a large number of negative samples, CPM mines positive samples according to the cosine similarity computed in complementary modalities. Differently in our approach, the CMD re-formulates cross-modal reinforcement as a Mutual Distillation (MD) process, with similarity distribution that models the modality-specific knowledge. To further alleviate the interference of numerous similar samples and facilitate better context modeling, IMD with Dynamic Neighbors Aggregation (DNA) is introduced for cluster-level contrastive learning and cross-level knowledge exchange.}
	\label{fig:overview}
\end{figure*}

Despite inspiring improvements, we argue that there are two factors limiting the existing contrastive learning based methods in this field.
On the one hand, integrating complementary information from different modalities has been validated to be significant for high-performance 3D human action recognition approaches~\citep{cheng2020skeleton,transvgpp,liang2019three,2sagcn2019cvpr,9234715}, but
\textit{cross-modal interactive learning is largely overlooked} in the self-supervised representation learning algorithms. Ideally, action prediction should be consistent across skeleton modalities, since they share the same high-level semantics. But in reality, the learned representation in each modality inevitably contains inherent bias, as they observe data in different "views". On the other hand, the conventional contrastive learning pipeline \textit{unanimously regards skeletons from different videos as negative samples}. Given the limited action categories, it would be unreasonable to just ignore potential similar instances, since they may belong to the same category. Ignoring this problem may lead to models struggling with similar instance discrimination thus without sufficient learning of class-level invariant features.

To tackle the aforementioned problems and facilitate better 3D action pre-training, we propose the Inter- and Intra-modal Mutual Distillation (I$^2$MD) framework. In I$^2$MD, the mutual distillation strategy is designed as a general bidirectional knowledge transfer module, which can be applied both between different skeleton modalities for cross-modal reinforcement and between the two branches within each modality for cross-level knowledge exchange.

By consolidating the idea of leveraging cross-modal reinforcement, we first introduce Cross-modal Mutual Distillation (CMD), where the idea underlying cross-modal positive mining (as shown in Fig.~\ref{fig:overview} (a)) is extended and re-formulated into a more general Mutual Distillation (MD) \citep{hinton2015distilling} mechanism. As illustrated in Fig.~\ref{fig:overview} (b), the neighboring similarity distribution is first extracted from each skeleton modality. It describes the relationship of the sample embedding with respect to its nearest neighbors in the customized feature space. Based on such relational information, bidirectional knowledge distillation is performed between every two modalities via explicit distribution consistency constraints. Compared to individual features \citep{hinton2015distilling} or logits \citep{romero2014fitnets}, the similarity distribution adopted in MD is more competitive for modeling the knowledge learned by contrastive frameworks in different modalities. Moreover, we also mathematically prove that under extreme settings, the proposed MD degenerate to positive sample mining.

CMD enables each skeleton modality to receive knowledge from other perspectives that is previously under-explored by itself due to the modal bias so that each skeleton modality learns more comprehensive representation. However, it does not solve the aforementioned second problem, \emph{i.e.}, the interference of numerous similar samples. To this end, in I$^2$MD, we further propose the Intra-modal Mutual Distillation (IMD) strategy. Specifically, we first go beyond conventional self-augmented instance-level discrimination by introducing the Dynamic Neighbors Aggregation (DNA) mechanism. As shown in Fig.~\ref{fig:overview} (c), an additional Cluster-level Discrimination Branch (CDB) is instantiated in each modality during pre-training. It takes both the training sample and its neighboring similar instances as inputs. By aggregating the encoded embedding features of these highly correlated neighbors, the contrastive learner exhibits better exploration of their shared invariant semantics. Despite the better cluster-level discrimination capability, the obtained features cannot be directly applied to the downstream tasks, as the neighboring information is no longer available after pre-training. An appropriate strategy is required to further transfer the learned representation to the original Instance-level Discrimination Branch (IDB). In our approach, this is achieved by reapplying the mutual distillation strategy between the two branches within each skeleton modality.

With the proposed I$^2$MD, the solutions to the two existing problems are naturally integrated into a unified framework, bridging the gap between conventional contrastive instance discrimination and 3D action representation learning. To validate the merits of our approach, we conduct extensive experiments on three prevalent datasets: NTU RGB+D 60 \citep{shahroudy2016ntu}, NTU RGB+D 120 \citep{liu2020ntu}, and PKU-MMD II \citep{liu2017pku}. Our approach shows outstanding results on all of them under all the four popular evaluation protocols and sets a series of new state-of-the-art records.

In summary, we make the following contributions:
\begin{itemize}
	\item In this work, we propose the Inter- and Intra-modal Mutual Distillation (I$^2$MD) framework for 3D action representation learning. In I$^2$MD, the Cross-modal Mutual Distillation (CMD) strategy is first introduced, which re-formulates cross-modal reinforcement as a bidirectional knowledge distillation process, \emph{i.e.}, mutual distillation.
	\item To alleviate the interference of numerous similar samples and exploit their underlying contexts, we further propose the Intra-modal Mutual Distillation (IMD) strategy. In IMD, the Dynamic Neighbors Aggregation (DNA) mechanism is introduced for cluster-level discrimination and the mutual distillation mechanism is again applied for cross-level knowledge exchange.
	\item The extensive experiments carried out on three prevalent benchmarks show that our I$^2$MD exhibits outstanding performance and sets several new state-of-the-art records.
\end{itemize}

This paper is an extension of our previous conference version \citep{mao2022cmd}. We have made significant improvements to the initial work. First, to alleviate the interference of numerous similar samples and explore the contextual information brought by the nearest neighbors, we introduce the Dynamic Neighbors Aggregation (DNA) strategy for cluster-level discriminative representation learning. The additional cross-attention module in the transformer architecture is ready to establish neighboring feature aggregation. Second, based on DNA, we extend the Cross-modal Mutual Distillation (CMD) in the preliminary version into a unified Inter- and Intra-modal Mutual Distillation (I$^2$MD) framework, with the mutual distillation mechanism abstracted for both inter- and intra-modal reinforcement. Third, we incorporate substantial new experimental results in this paper, including SOTA comparison, ablative study, and qualitative analysis. Our source code will be made publicly available at: \url{https://github.com/maoyunyao/I2MD}.

\section{Related Work}
In this section, we perform a literature review on self-supervised representation learning, self-supervised 3D action recognition, and similarity-based knowledge distillation, respectively.

\subsection{Self-supervised Representation Learning}
According to the different pretexts used, recent self-supervised learning approaches can be roughly divided into two categories: generative and contrastive \citep{liu2021self}. Generative methods \citep{ballard1987modular,he2022masked,van2017neural} try to reconstruct the original input to learn meaningful latent representation, including autoregressive models \citep{yang2019xlnet,van2016pixel}, flow-based models \citep{dinh2014nice,dinh2016density}, autoencoders \citep{ballard1987modular,he2022masked,van2017neural}, \emph{etc.} Contrastive learning \citep{chen2020simple,he2020momentum,van2018representation} aims to learn feature representation via instance discrimination. It pulls positive pairs closer and pushes negative pairs away. Since no labels are available during self-supervised contrastive learning, two different augmented versions of the same sample are treated as a positive pair, and samples from different instances are considered to be negative. In MoCo \citep{he2020momentum} and MoCo v2 \citep{chen2020improved}, the negative samples are taken from previous batches and stored in a queue-based memory bank. In contrast, SimCLR \citep{chen2020simple} and MoCo v3 \citep{chen2021empirical} rely on a larger batch size to provide sufficient negative samples. Despite achieving outstanding performance on benchmarks like ImageNet \citep{imagenet2009A}, these standard contrastive learning methods do not consider the potential positive samples in the memory bank in case of limited categories, let alone the complementary properties between different skeleton modalities in 3D action representation learning. The goal of our proposed I$^2$MD is to tackle these problems and thus facilitate better contrastive 3D action representation learning.

\subsection{Self-supervised 3D Action Recognition}
Many previous works have been proposed to perform self-supervised 3D action representation learning. In LongT GAN \citep{zheng2018unsupervised}, an autoencoder-based model along with an additional adversarial training strategy are proposed. Following the generative paradigm, it learns latent representation via original skeleton sequence reconstruction. Similarly, P\&C \citep{su2020predict} trains an encoder-decoder network to both predict and cluster skeleton sequences. To learn features that are more robust and discriminative, the authors also propose strategies to weaken the decoder, laying more burdens on the encoder. Different from the aforementioned methods that adopt a single reconstruction task, MS$^2$L \citep{lin_ms2l_mm} integrates multiple pretext tasks, including motion prediction, jigsaw puzzle recognition, and contrastive learning, to learn better 3D action representation.

In recent attempts \citep{li20213d,rao2021augmented,thoker2021skeleton,guo2022aimclr,rao2023hierarchical}, momentum encoder-based contrastive learning is introduced and better performance is achieved. Among them, CrosSCLR \citep{li20213d} is the first to perform cross-modal knowledge mining. Specifically, it finds potential positive anchors and re-weights training samples with the contrastive context from complementary skeleton modalities. However, the positive sample mining in CrosSCLR requires reliable preliminary knowledge, the representation in each modality needs to be optimized independently in advance, leading to a sophisticated two-stage training process. Moreover, the contrastive context, defined as the similarity between the positive query and negative embedding features, is treated as individual weights of samples in complementary modalities to participate in the optimization process. Such implicit knowledge exchange lacks a holistic grasp of the rich contextual information. Differently, in our approach, a general mutual distillation mechanism is introduced to perform inter- and intra-modal interactive learning. Instead of heuristic positive mining or sample re-weighting, in I$^2$MD, the contrastive context is treated as a whole to describe the modality-specific knowledge from a probabilistic perspective, avoiding tedious two-stage pre-training. Besides, cross-modal positive mining can be regarded as a special case of the proposed mutual distillation strategy performed under the inter-modal scenario.

\subsection{Similarity-based Knowledge Distillation}
Pairwise similarity has been shown to be useful information in relational knowledge distillation \citep{park2019relational,peng2019correlation,tung2019similarity}. In PKT \citep{passalis2018learning}, CompRess \citep{abbasi2020compress}, and SEED \citep{fang2021seed}, similarities of each sample with respect to a set of random anchors are converted into a probability distribution, which models the structural information of the data representation. After that, knowledge distillation is performed by training the student to mimic the probability distribution of the pre-trained and fixed teacher. Recently, contextual similarity information has also shown great potential in many areas including visual re-ranking \citep{ouyang2021contextual}, asymmetric image retrieval \citep{wu2022contextual}, and representation learning \citep{ALBEF,Tejankar_2021_ICCV}. In our approach, we also find that the pairwise similarity-based representation is naturally suitable for modeling the modality-specific 3D action knowledge learned with contrastive learning frameworks. Our proposed framework is partially inspired by the aforementioned remarkable works. Different from \citep{passalis2018learning,abbasi2020compress,fang2021seed} that try to distill the knowledge from a pre-trained and fixed teacher to a smaller student, in I$^2$MD, the knowledge is continuously updated and bidirectionally distilled. Different from the momentum distillation introduced in \citep{ALBEF} that tries to help align the image and text representations before fusing them through cross-modal attention, the mutual distillation in our approach is designed to answer the question of how to transfer the biased knowledge between complementary skeleton modalities or representation levels for more comprehensive 3D action pre-training.

\section{Our Approach}

\begin{figure*}[t!]
    \centering
    \includegraphics[width=1.0\linewidth]{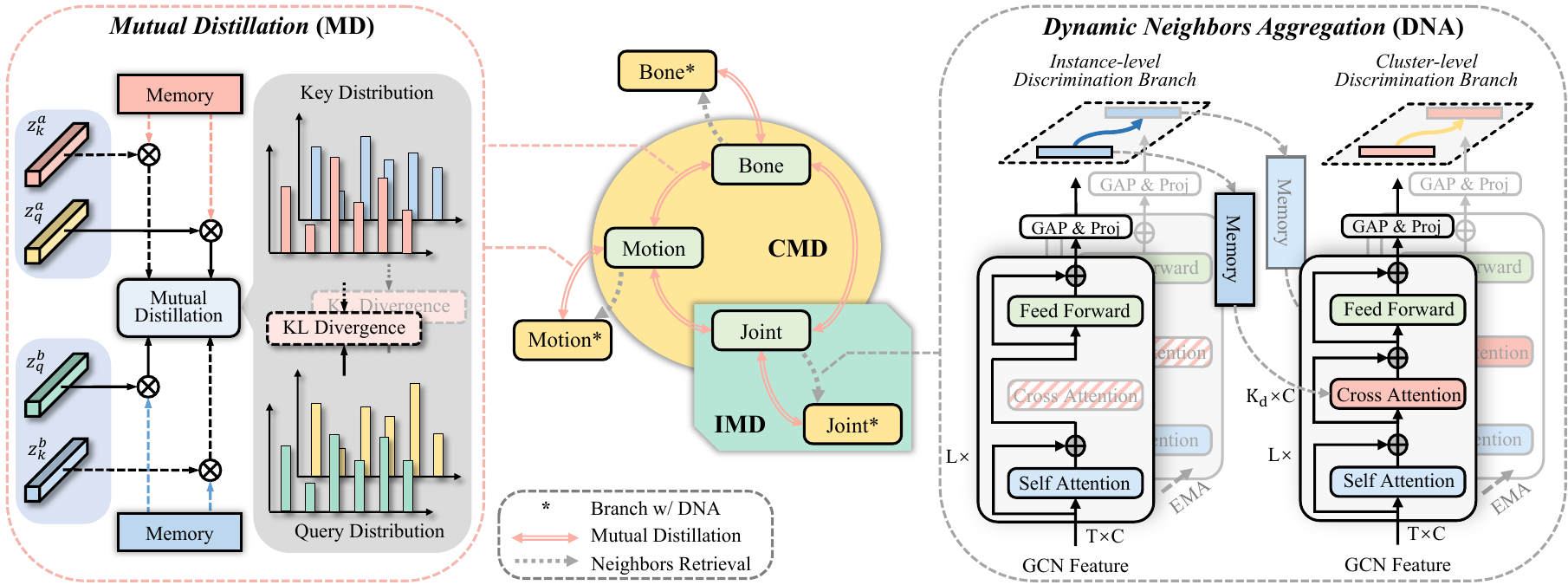}
    \caption{An illustration of our proposed Inter- and Intra-modal Mutual Distillation (I$^2$MD) framework. It consists of two components, namely Cross-modal Mutual Distillation (CMD) and Intra-modal Mutual Distillation (IMD). In CMD, the Mutual Distillation (MD) mechanism is applied between different skeleton modalities for cross-modal interactive learning. In IMD, the Dynamic Neighbors Aggregation (DNA) mechanism is first introduced, where the neighboring features are retrieved from memory and aggregated by the cluster-level discrimination branch, forming local cluster-level discriminative representation learning. Then the mutual distillation mechanism is again utilized between the two branches within each skeleton modality for cross-level knowledge exchange.}\label{fig:i2md}
\end{figure*}

In this section, we introduce the proposed Inter- and Intra-modal Mutual Distillation (I$^2$MD) framework in detail. We first review the background knowledge and the underlying limitations of the standard contrastive learning algorithm in Section~\ref{sec:background}. Then in Section~\ref{subsec:cmd}, we elaborate Cross-modal Mutual Distillation (CMD), where the concept of Mutual Distillation (MD) is first introduced for cross-modal reinforcement. After that, in Section~\ref{subsec:iimd}, we introduce Intra-modal Mutual Distillation (IMD), where the Dynamic Neighbors Aggregation (DNA) mechanism is equipped on an additional branch for cluster-level discrimination. The mutual distillation strategy, as a general knowledge transfer scheme, is again applied between the two branches within each skeleton modality for cross-level knowledge exchange. Figure \ref{fig:i2md} illustrates the overall architecture of our proposed I$^2$MD framework.

\subsection{Background: Contrastive Learning}
\label{sec:background}

In this section, we revisit the single-modal contrastive learning as the background of our approach, which has been widely adopted in many tasks like image/video recognition \citep {imagenet2009A,kuehne2011hmdb,soomro2012ucf101} and correspondence learning \citep{wang2021contrastive}. In self-supervised 3D action recognition, previous works like AS-CAL \citep{rao2021augmented},  CrosSCLR \citep{li20213d}, ISC \citep{thoker2021skeleton}, and AimCLR \citep{guo2022aimclr} also take the contrastive method MoCo v2 \citep{chen2020improved} as their baseline.

Given a single-modal skeleton sequence $x$, we first perform data augmentation to obtain two different views $x_q$ and $x_k$ (query and key). Then, two encoders are adopted to map the positive pair $x_q$ and $x_k$ into feature embeddings $z_q = E_q(x_q, \theta_q)$ and $z_k = E_k(x_k, \theta_k)$, where $E_q$ and $E_k$ denote query encoder and key encoder, respectively. $\theta_q$ and $\theta_k$ are the learnable parameters of the two encoders. Note that in MoCo v2, the key encoder is not trained by gradient descent but the momentum-updated version of the query encoder:
$\theta_k \gets \alpha \theta_k + (1-\alpha)\theta_q$, where $\alpha$ is a momentum coefficient that controls the updating speed. During self-supervised pre-training, the noise contrastive estimation loss InfoNCE \citep{van2018representation} is used to perform instance discrimination, which is computed as follows:
\begin{equation}
	\begin{aligned}
		\mathcal{L}_{\text{SCL}}=
		-\log \frac{\exp (z_q ^\top z_k / \tau_c)}{\exp (z_q ^\top z_k / \tau_c)+\sum_{i=1}^{N} \exp \left(z_q ^\top m_{i} / \tau_c \right)},
	\end{aligned}
\end{equation}
where $\tau_c$ is a temperature hyper-parameter \citep{hinton2015distilling} that scales the distribution of instances and $m_i$ is the key embedding of negative sample. $N$ is the size of a queue-based memory bank $\mathbf{M}$ where all the negative key embeddings are stored. After the training of the current mini-batch, $z_k$ is enqueued as a new negative key embedding and the oldest embedding in the memory bank are dequeued.

Under the supervision of the InfoNCE loss, the encoder is forced to learn representation that is invariant to data augmentations, thereby focusing on semantic information shared between positive pairs. Nevertheless, the learned representation is often modally biased, making it difficult to account for all data characteristics. Though it can be partially alleviated by test-time ensembling, several times the running overhead will be introduced. And the inherent limitations of the representation in each skeleton modality still exist. Moreover, since there is no category-level supervision available, all the data points from different skeleton sequences are considered negative, which is unreasonable, especially when the number of action categories is limited.

\subsection{Cross-modal Mutual Distillation (CMD)}
\label{subsec:cmd}
By consolidating the idea of leveraging complementary information from cross-modal inputs to improve 3D action representation learning, we design the cross-modal mutual distillation strategy. We first introduce the Mutual Distillation (MD) mechanism in Section~\ref{subsubsec:md} and then explain how it is used for cross-modal reinforcement in Section~\ref{subsubsec:md-in-cmd}. Finally, we derive the relation between mutual distillation and mutual positive sample mining in Section~\ref{subsubsec:relation}.

\subsubsection{Mutual Distillation (MD)}
\label{subsubsec:md}
The mutual distillation mechanism is designed for knowledge exchange between two contrastive learning entities.

\parsection{Knowledge Modeling}
To perform knowledge distillation, we first need to model the learned knowledge in a proper way. It needs to take advantage of the existing contrastive framework to avoid introducing excessive computational overhead. Moreover, the two entities in mutual distillation may come from different modalities or levels of representation, conventional methods that rely on individual features/logits are no longer applicable. 

Inspired by recent relational knowledge distillation works \citep{park2019relational,peng2019correlation,tung2019similarity}, we utilize the pairwise relationship between samples for knowledge modeling. Given an embedding $z$ and a set of anchors $\{n_i\}_{i = 1,2,\cdots, K_c}$, we compute the similarities between them as $\text{sim}(z, n_i) = z^\top n_i, i = 1,2, \cdots, K_c.$ In MoCo v2 \citep{chen2020improved}, there are a handful of negative embedding features stored in the memory. We can easily obtain the required anchors without additional model inference. Note that if all the negative embedding features are used as anchors, the set $\{z^\top m_i\}_{i = 1,2, \cdots, N}$ is exactly the contrastive context defined in \citep{li20213d}. In our approach, we select the top $K_c$ nearest neighbors of $z$ as the anchors. The resulting pairwise similarities are further converted into probability distributions with a temperature $\tau$ :
\begin{equation}
	\label{eq:distribution}
	\begin{aligned}
		p_i(z, \tau) = \frac{\exp(z ^\top n_i  / \tau )}{\sum_{j=1}^{K_c} \exp(z ^\top n_j / \tau )}, i = 1,2, \cdots, K_c.
	\end{aligned}
\end{equation}
The obtained $ \boldsymbol{p}(z, \tau) = \{p_i(z, \tau)\}_{i = 1,2,\cdots, K_c}$ describes the distribution characteristic around the embedding $z$ in the customized feature space of each contrastive learning entity.

\parsection{Knowledge Distillation}
Based on the aforementioned probability distributions, an intuitive way to perform knowledge distillation is to directly establish consistency constraints in between. Different from previous knowledge distillation approaches that transfer the knowledge of a fixed and well-trained teacher to the student, in our approach, the knowledge is continuously updated during pre-training and bidirectionally distilled between two contrastive learning entities.

To this end, based on the contrastive framework, we make two customized designs: \textbf{i)} Different embedding features are used for the teacher and student. As shown in Fig.~\ref{fig:i2md}, in MoCo v2 \citep{chen2020improved}, two augmented views of the same sample are encoded into query $z_q$ and key $z_k$, respectively. In our approach, the key distribution obtained in one entity is used to guide the learning of query distribution in the other entity, so that knowledge is transferred accordingly. Specifically, for the key embedding $z_k^a$ from entity A and the query embedding $z_q^b$ from entity B, we select the top $K_c$ nearest neighbors of $z_k^a$ as anchors and compute the similarity distributions as $\boldsymbol{p}(z_q^b, \tau)$ and $\boldsymbol{p}(z_k^a, \tau)$ according to Eq.~\ref{eq:distribution}. Knowledge distillation from entity A to entity B is performed by minimizing the following KL divergence:
\begin{equation}
	\begin{aligned}
		\mathrm{KL}\big(\boldsymbol{p}(z_k^a, \tau) || \boldsymbol{p}(z_q^b, \tau)\big) = \sum_{i=1}^{K_c} p_i(z_k^a, \tau) \cdot \log \frac{ p_i(z_k^a, \tau)}{p_i(z_q^b, \tau)}.
	\end{aligned}
\end{equation}
Since the key encoder is not trained with gradient, the teacher is not affected during unidirectional knowledge distillation. Moreover, the momentum-updated key encoder provides more stable knowledge for the student to learn. \textbf{ii)} Asymmetric temperatures $\tau_t$ and $\tau_s$ are employed for teacher and student, respectively. Considering that there is no intuitive teacher-student relationship in between, a smaller temperature is applied for the teacher to emphasize the high-confidence information, as in \citep{Tejankar_2021_ICCV}.

Since knowledge distillation works bidirectionally, given two entities A and B, the loss function for mutual distillation is formulated as follows:
\begin{equation}
    \label{eq:loss_md}
    \begin{aligned}
        \mathcal{L}_{\text{MD}}^{a \leftrightarrow b}
        = \, &\mathrm{KL}(\boldsymbol{p}(z_k^a, \tau_t) || \boldsymbol{p}(z_q^b, \tau_s)) \\
        + \, &\mathrm{KL}(\boldsymbol{p}(z_k^b, \tau_t) || \boldsymbol{p}(z_q^a, \tau_s)).
    \end{aligned}
\end{equation}

\begin{figure}[t!]
    \centering
    \includegraphics[width=1.0\linewidth]{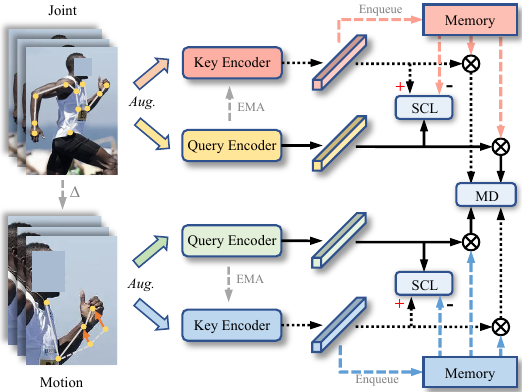}
    \caption{The overall pipeline of CMD. Given multiple skeleton modalities (\emph{e.g.} joint and motion) as input, Single-modal Contrastive Learning (SCL) is firstly applied to perform self-supervised learning within each modality. Meanwhile, the proposed Mutual Distillation (MD) models and bidirectionally transfers the customized knowledge between modalities.}\label{fig:CMD}
\end{figure}

\subsubsection{Cross-modal Reinforcement with MD}
\label{subsubsec:md-in-cmd}
Fig.~\ref{fig:CMD} shows the pipeline when the mutual distillation mechanism is applied between joint and motion modalities for cross-modal reinforcement. While single-modal contrastive learning (SCL) is performed within each skeleton modality, the proposed mutual distillation mechanism models the learned knowledge and transfers it between modalities. This enables each modality to receive knowledge from other perspectives, thereby alleviating the modal bias of the learned representation. Given three skeleton modalities joint, motion, and bone, the loss function for cross-modal mutual distillation is formulated as follows:
\begin{equation}
    \label{eq:loss_cmd}
    \begin{aligned}
        \mathcal{L}_{\text{CMD}} = \mathcal{L}_{\text{MD}}^{j \leftrightarrow m} + \mathcal{L}_{\text{MD}}^{m \leftrightarrow b} + \mathcal{L}_{\text{MD}}^{b \leftrightarrow j},
    \end{aligned}
\end{equation}
where $j$, $m$, and $b$ denote joint, motion, and bone modality, respectively.

\subsubsection{Relationship with Positive Mining}
\label{subsubsec:relation}
We discuss the relationship between the proposed mutual distillation and positive sample mining. Setting temperature $\tau_t = 0$ and $K_c=N$, the key distribution $\boldsymbol{p}(z_k^a, \tau_t)$ and $\boldsymbol{p}(z_k^b, \tau_t)$ in Eq.~\ref{eq:loss_md} will be one-hot vectors with the only $1$ at index $u$ and $v$, respectively. Thus the loss function will be like:
\begin{equation}
	\label{eq:relation}
	\begin{aligned}
		\mathcal{L}_{\text{MD}}^{a \leftrightarrow b}
        &= \mathrm{KL}(\boldsymbol{p}(z_k^a, 0) || \boldsymbol{p}(z_q^b, \tau_s))
         + \mathrm{KL}(\boldsymbol{p}(z_k^b, 0) || \boldsymbol{p}(z_q^a, \tau_s)) \\
		&= 1 \cdot\log \frac{1}{p_u(z_q^b, \tau_s)} + 1 \cdot\log \frac{1}{p_v(z_q^a, \tau_s)} \\
		&= - \log \frac{\exp({z_q^b} ^\top m_u^b  / \tau_s )}{\sum_{j=1}^{N} \exp({z_q^b} ^\top m_j^b / \tau_s )} \\
        & \,\,\,\,\,\,\, - \log \frac{\exp({z_q^a} ^\top m_v^a  / \tau_s )}{\sum_{j=1}^{N} \exp({z_q^a} ^\top m_j^a / \tau_s )}.
	\end{aligned}
\end{equation}
We can find that the loss $\mathcal{L}_{\text{MD}}^{a \leftrightarrow b}$ is essentially doing contrastive learning in entity B and A with the positive sample $m_u^b$ and $m_v^a$ mined by entity A and B. Thus we draw a conclusion that under extreme settings, the proposed mutual distillation degenerates to mutual positive sample mining.

\subsection{Intra-modal Mutual Distillation (IMD)}
\label{subsec:iimd}
To further alleviate the interference of numerous similar samples and better explore the contextual information, in each skeleton modality, we further propose the intra-modal mutual distillation strategy. We first introduce the Dynamic Neighbors Aggregation (DNA) mechanism in Section~\ref{subsubsec:dna}, which enables local cluster-level contrastive representation learning. Then in Section~\ref{subsubsec:dna-for-clke}, we show how the mutual distillation mechanism is applied within each skeleton modality for cross-level knowledge exchange.

\subsubsection{Dynamic Neighbors Aggregation (DNA)}
\label{subsubsec:dna}
The dynamic neighbors aggregation mechanism enables the network to adaptively aggregate the features from the nearest neighbors of the training sample, producing cluster-level representation.

\parsection{Network Architecture}
To facilitate the idea of dynamic neighbors aggregation, we re-design the skeleton feature extractor. Specifically, we adopt a GCN-Transformer cascaded network, where a three-layer Graph Convolutional Neural Network is utilized before the Transformer to extract frame-level features $f \in R^{T \times C}$ according to the internal topology of the human skeletons. The obtained features, as $T$ temporal tokens, are then fed into a Transformer consisting of a stack of $L$ identical layers, where each layer consists of a self-attention block, a cross-attention block (optional, for dynamic neighbors aggregation), and a feed-forward block. The output is further pooled along the temporal dimension to obtain the global hidden embedding $h \in R^C$, which can be processed by a fully connected layer to perform 3D action classification.

\parsection{Cluster-level Discrimination}
As shown in Fig.~\ref{fig:i2md}, in each skeleton modality, we additionally instantiate a weight-sharing cluster-level discrimination branch (CDB in short, on the right) parallel to the original instance-level discrimination branch (IDB in short, on the left), which is equipped with an additional cross-attention block in each layer for neighboring feature aggregation. During pre-training, the query $z_q \in R^{C'}$ and key $z_k \in R^{C'}$ from IDB are used to search the nearest neighbors of the training sample from the memory bank. The corresponding hidden embedding of the top $K_d$ nearest neighbors are retrieved and are concatenated as the value input $H_q \in R^{K_d \times C}$ and $H_k \in R^{K_d \times C}$ of the feature aggregation module, \emph{i.e.}, the cross-attention blocks in CDB. The output of CDB $z_q^* \in R^{C'}$ and $z_k^* \in R^{C'}$ thus integrate local contextual information around the training sample in the feature space. Based on this, we construct local cluster-level contrastive learning as follows:
\begin{equation}
	\begin{aligned}
		\mathcal{L}_{\text{SCL}}^{*}=
		-\log \frac{\exp (z_q^{* \top} z_k^{*} / \tau_c)}{\exp (z_q^{*\top} z_k^{*} / \tau_c)+\sum_{i=1}^{N} \exp \left(z_q^{*\top} m_{i}^{*} / \tau_c \right)},
	\end{aligned}
\end{equation}
where $m_i^{*}$ is the key embedding features of negative samples saved in previous batches, as discussed in Section~\ref{sec:background}. To facilitate the retrieval of hidden embedding, we need to store them into the memory along with the corresponding keys, forming key-value pairs.

\subsubsection{Cross-level Knowledge Exchange}
\label{subsubsec:dna-for-clke}

Although CDB learns cluster-aware representation, it cannot be directly applied to the downstream tasks, as the neighboring information is no longer available after pre-training. To this end, we consider establishing knowledge transfer between the two branches so that the original IDB performs better representation learning under the guidance of CDB. Instead of simply applying cross-branch contrastive learning, we consider an alternative way, extending the idea of mutual distillation in CMD into the intra-modal scenario. Specifically, in each skeleton modality, we employ the following loss function between the IDB and CDB:
\begin{equation}
	\label{eq:loss_imd}
	\begin{aligned}
		\mathcal{L}_{\text{IMD}} = \mathcal{L}_{\text{MD}}^{i \leftrightarrow c}
		=\, &\mathrm{KL}\big(\boldsymbol{p}(z_k^{*}, \tau_t) || \boldsymbol{p}(z_q, \tau_s)\big) \\
		+\, &\mathrm{KL}\big(\boldsymbol{p}(z_k, \tau_t) || \boldsymbol{p}(z_q^{*}, \tau_s)\big),
	\end{aligned}
\end{equation}
where the superscripts $i$ and $c$ denote the instance-level discrimination branch and cluster-level discrimination branch, respectively. We can find that $\mathcal{L}_{\text{IMD}}$ and $\mathcal{L}_{\text{CMD}}$ exhibits the same form, the only difference being the source and destination of the knowledge to be distilled. The loss function inside each skeleton modality is then formulated as follows:

\begin{equation}
	\begin{aligned}
		\mathcal{L}_{\text{intra}} &= \mathcal{L}_{\text{SCL}} + \mathcal{L}_{\text{SCL}}^{*} + \mathcal{L}_{\text{IMD}}.
	\end{aligned}
\end{equation}

\subsection{Final Pre-training Objective}
The final pre-training objective of the proposed I$^2$MD is the combination of $\mathcal{L}_{\text{intra}}$ and $\mathcal{L}_{\text{CMD}}$, which is formulated as follows:
\begin{equation}
	\label{eq:loss_final}
	\begin{aligned}
		\mathcal{L}_{\text{final}} = \mathcal{L}_{\text{intra}}^j + \mathcal{L}_{\text{intra}}^m + \mathcal{L}_{\text{intra}}^b + \lambda_1 \cdot \mathcal{L}_{\text{CMD}},
	\end{aligned}
\end{equation}
where $\lambda_1$ is the constant coefficient that balances the training loss. The superscripts $j$, $m$, and $b$ denote joint, motion, and bone modality, respectively.

\section{Experiments}

\subsection{Implementation Details}
\subsubsection{Network Architecture} In our preliminary CMD framework, we adopt a 3-layer Bidirectional GRU (BiGRU) as the base encoder, which has a hidden dimension of 1024. Before the encoder, we additionally add a Batch Normalization \citep{ioffe2015batch} layer to stabilize the training process. In the advanced I$^2$MD framework, the encoder is replaced by the GCN-Transformer cascaded network introduced in Section~\ref{subsubsec:dna}. We adopt the first three layers of ST-GCN \citep{yan2018spatial} as the frame-level feature extractor, which extracts features with a dimension of $3200$ for each 3D skeleton frame. For the Transformer, the number of layers $L$ is 6, the embedding dimension is 768, the head number of the multi-head attention module is 12, the hidden dimension in FFN is 3072. Between the GCN and the Transformer, we additionally add a linear layer to reduce the GCN feature channel from 3200 to 768.

\subsubsection{Self-supervised Pre-training} During pre-training, we adopt MoCo v2 \citep{chen2020improved} to perform single-modal contrastive learning, where an MLP head is attached to each encoder to map the representation into a 128-dimensional embedding, which is L2-normalized for further computation. The temperature hyper-parameter in the InfoNCE \citep{van2018representation} loss is 0.07. In the proposed mutual distillation, the temperatures for the teacher and student are set to 0.05 and 0.1, respectively. The number of neighbors $K_c$ and $K_d$ for CMD and DNA are set to 8192 and 64, respectively. The constant coefficient $\lambda_1$ in Eq.~\ref{eq:loss_final} is set to 0.5. The SGD optimizer is employed with a momentum of 0.9 and a weight decay of 0.0001. The batch size is set to 256 and the initial learning rate is 0.01. For NTU RGB+D 60 \citep{shahroudy2016ntu} and NTU RGB+D 120 \citep{liu2020ntu} datasets, the model is trained for 500 epochs, the learning rate is reduced to 0.001 after 400 epochs, and the size of the memory bank $N$ is 16384. For PKU-MMD II \citep{liu2017pku} dataset, the total epochs are increased to 1000, and the learning rate drops at epoch 800. We adopt the same skeleton augmentations as ISC \citep{thoker2021skeleton}.

\subsection{Datasets and Metrics}
\subsubsection{NTU RGB+D 60 \citep{shahroudy2016ntu}} NTU-RGB+D 60 (NTU-60) is a large-scale multi-modality action recognition dataset that is captured by three Kinect V2 cameras. It contains 60 action categories and 56,880 sequences. The actions are performed by 40 different subjects (actors). In this paper, we adopt its 3D skeleton data for experiments. Specifically, each human skeleton contains 25 body joints, and each joint is represented as 3D coordinates. Two evaluation protocols are recommended by the authors: cross-subject (x-sub) and cross-view (x-view). For x-sub, action sequences performed by half of the 40 subjects are used for training and the rest for testing. For x-view, the training samples are captured by cameras 2 and 3 and the test samples are from camera 1.

\subsubsection{NTU RGB+D 120 \citep{liu2020ntu}} Compared with NTU-60, NTU-RGB+D 120 (NTU-120) extends the action categories from 60 to 120, with 114,480 skeleton sequences in total. The number of subjects is also increased from 40 to 106. Moreover, a new evaluation protocol named cross-setup (x-set) is proposed as a substitute for x-view. Specifically, the sequences are divided into 32 different setups according to the camera distances and background, with half of the 32 setups (even-numbered) used for training and the rest for testing.

\subsubsection{PKU-MMD \citep{liu2017pku}} PKU-MMD is a new benchmark for multi-modality 3D human action detection. It can also be used for action recognition tasks \citep{lin_ms2l_mm}. PKU-MMD has two phases, where Phase II is extremely challenging since more noise is introduced by large view variation. 
In this work, we evaluate the proposed method on Phase II (PKU-II) under the widely used cross-subject evaluation protocol, with 5,332 skeleton sequences for training and 1,613 for testing.

\subsubsection{Evaluation Metrics} We report the top-1 accuracy.

\subsection{Comparison with State-of-the-art Methods}

In the section, the learned representation is utilized for 3D action classification under a variety of evaluation protocols. We compare our proposed CMD (preliminary version) and I$^2$MD  (advanced version) with previous state-of-the-art methods. During evaluation, we only take a single skeleton modality (joint) as input by default, which is consistent with previous arts \citep{lin_ms2l_mm,su2020predict,thoker2021skeleton}. Modality ensembling can improve performance but is time-consuming.

\begin{table*}[t!]
    \centering
        \setlength\tabcolsep{5.5pt}
        \caption{Performance comparison on NTU-60, NTU-120, and PKU-II in terms of the linear evaluation protocol.}\label{tab:sota_comp_linear}
        \begin{tabular*}{0.8\linewidth}{lcccccc}
            \toprule[1.2pt]
            \multirow{2}{*}{Method}  & \multirow{2}{*}{Venue} &\multicolumn{2}{c}{\textbf{NTU-60}}	& \multicolumn{2}{c}{\textbf{NTU-120}} & \textbf{PKU-II}\\
            \cmidrule[0.8pt](lr){3-4} \cmidrule[0.8pt](lr){5-6} \cmidrule[0.8pt](lr){7-7}
                                                     & & x-sub & x-view & x-sub & x-set & x-sub\\
            \midrule[0.8pt]
            LongT GAN \citep{zheng2018unsupervised}  & AAAI'2018 & 39.1  & 48.1   & -     & -     & 26.0 \\
            MS$^2$L \citep{lin_ms2l_mm}              & ACM MM'2020 & 52.6  & -      & -     & -     & 27.6 \\
            P\&C \citep{su2020predict}		         & CVPR'2020 & 50.7  & 76.3   & 42.7  & 41.7  & 25.5 \\
            AS-CAL \citep{rao2021augmented} 		 & Inf. Sci.'2021 & 58.5  & 64.8   & 48.6  & 49.2  & -    \\
            PCRP \citep{9623511} 			         & TMM'2023 & 56.5  & 66.6   & 44.8  & 46.9  & -    \\
            ST-CL \citep{9612062} 			         & TMM'2023 & 68.1  & 69.4   & 54.2  & 55.6  & -    \\
            SeBiReNet \citep{nie2020unsupervised}    & ECCV'2020 & -     & 79.7   & -     & -     & -    \\
            AimCLR \citep{guo2022aimclr}	         & AAAI'2022 & 74.3  & 79.7   & -     & -     & -    \\
            ISC \citep{thoker2021skeleton}		     & ACM MM'2021 & 76.3  & 85.2   & 67.1  & 67.9  & 36.0 \\
            GL-Transformer \citep{gl_transformer}    & ECCV’2022 & 76.3  & 83.8   & 66.0  & 68.7  & -    \\
            CrosSCLR-B \citep{li20213d}	             & CVPR’2021 & 77.3  & 85.1   & 67.1  & 68.6  & 41.9 \\
            CPM \citep{zhang2022contrastive}	     & ECCV’2022 & 78.7  & 84.9   & 68.7  & 69.6  & 48.3 \\
            3s-HiCLR \citep{zhang2023hierarchical}   & AAAI'2023 & 80.4  & 85.5   & 70.0  & 70.4  & -    \\
            ActCLR \citep{Lin_2023_CVPR}             & CVPR’2023 & 80.9  & 86.7   & 69.0  & 70.5  & -    \\
            \midrule[0.8pt]
            \textbf{CMD (Ours)}                      & ECCV’2022 & 79.4  & 86.9	  & 70.3  & 71.5  & 43.0 \\
            \textbf{I$^2$MD (Ours)}                  & & \textbf{83.4}  & \textbf{88.0}	& \textbf{73.1} & \textbf{74.1} & \textbf{49.0} \\
            \bottomrule[1.2pt]
        \end{tabular*}
\end{table*}

\begin{table*}[t!]
    \centering
    \setlength\tabcolsep{10pt}
    \caption{Performance comparison on NTU-60 and NTU-120 in terms of the KNN evaluation protocol.}\label{tab:sota_comp_knn}
    \begin{tabular*}{0.8\linewidth}{lccccc}
        \toprule[1.2pt]
        \multirow{2}{*}{Method}   	& \multirow{2}{*}{Venue}	  & \multicolumn{2}{c}{\textbf{NTU-60}}	& \multicolumn{2}{c}{\textbf{NTU-120}}\\
        \cmidrule[0.8pt](lr){3-4} \cmidrule[0.8pt](lr){5-6}
        & & x-sub          & x-view  			& x-sub         	& x-set\\
        \midrule[0.8pt]
        LongT GAN \citep{zheng2018unsupervised}  & AAAI'2018 & 39.1           & 48.1   		& 31.5       & 35.5 \\
        P\&C \citep{su2020predict}               & CVPR'2020 & 50.7           & 76.3      	& 39.5       & 41.8 \\
        ISC \citep{thoker2021skeleton}		     & ACM MM'2021 & 62.5           & 82.6   		& 50.6       & 52.3 \\
        CrosSCLR-B \citep{li20213d}              & CVPR’2021 & 66.1           & 81.3   		& 52.5       & 54.9 \\
        HiCLR \citep{zhang2023hierarchical}      & AAAI'2023 & 67.3           & 75.3         & -          & -    \\
        \midrule[0.8pt]
        \textbf{CMD (Ours)}                      & ECCV’2022 & 70.6  & \textbf{85.4}	& 58.3 & 60.9 \\
        \textbf{I$^2$MD (Ours)}                  & & \textbf{75.9}  & 83.8	& \textbf{62.0} & \textbf{64.7} \\
        \bottomrule[1.2pt]
    \end{tabular*}
\end{table*}

\begin{table*}[t!]
    \centering
    \setlength\tabcolsep{9pt}
    \caption{Performance comparison on NTU-60 and NTU-120 in terms of the fine-tuned evaluation protocol.}\label{tab:sota_comp_finetune}
    \begin{tabular*}{0.8\linewidth}{lccccc}
        \toprule[1.2pt]
        \multirow{2}{*}{Method}   &\multirow{2}{*}{Venue}   & \multicolumn{2}{c}{\textbf{NTU-60}}	& \multicolumn{2}{c}{\textbf{NTU-120}}\\
        \cmidrule[0.8pt](lr){3-4} \cmidrule[0.8pt](lr){5-6}
        & & x-sub          & x-view  			& x-sub         	& x-set\\
        \midrule[0.8pt]
        TSRJI \citep{tsrji}                      & SIBGRAPI'2019 & 73.3  & 80.3  & 67.9  & 62.8  \\
        C-CNN + MTLN \citep{ke2017new}           & CVPR'2017 & 79.6  & 84.8  & -     & -     \\
        AimCLR \citep{guo2022aimclr}             & AAAI'2022 & 83.0  & 89.2  & 76.4  & 76.7  \\
        Colorization \citep{yang2021skeleton}    & ICCV'2021 & 84.2  & 93.1  & -     & -     \\
        CPM \citep{zhang2022contrastive}         & ECCV'2022 & 84.8  & 91.1  & 78.4  & 78.9  \\
        ActCLR \citep{Lin_2023_CVPR}             & CVPR’2023 & 85.8  & 91.2  & \textbf{79.4}  & \textbf{80.9}  \\
        \midrule[0.8pt]
        W/o pre-training            & ECCV'2022 & 80.1    & 91.2  & 71.1  & 74.1 \\
        \textbf{I$^2$MD (Ours)}     & & \textbf{86.5}  & \textbf{93.6}	& \textbf{79.1} & \textbf{80.3} \\
        \bottomrule[1.2pt]
    \end{tabular*}
\end{table*}

\subsubsection{Linear Evaluation Protocol} For linear evaluation protocol, we freeze the pre-trained encoder and add a learnable linear classifier after it. The classifier is trained on the corresponding training set for 80 epochs with a learning rate of 0.1 (reduced to 0.01 and 0.001 at epoch 50 and 70, respectively). We evaluate the proposed method on the NTU-60, NTU-120, and PKU-II datasets. As shown in Table~\ref{tab:sota_comp_linear}, we include the recently proposed CrosSCLR \citep{li20213d}, ISC \citep{thoker2021skeleton}, AimCLR \citep{guo2022aimclr}, CPM \citep{zhang2022contrastive}, HiCLR \citep{zhang2023hierarchical}, and ActCLR \citep{Lin_2023_CVPR} for comparison. Our CMD and I$^2$MD consistently outperform previous state-of-the-art methods. Note that ISC and CMD share the same BiGRU encoder, which is different from the ST-GCN \citep{yan2018spatial} encoder in CrosSCLR. For fair comparisons, we additionally train a variant of CrossSCLR with BiGRU as its base encoder (CrosSCLR-B in short). We can find that our method still outperforms it on all three datasets, which shows the superiority of the proposed cross-modal mutual distillation.

\begin{table*}[t!]
    \centering
    \setlength\tabcolsep{9pt}
    \caption{Performance comparison on NTU-60 in terms of the semi-supervised evaluation protocol. As in \citep{thoker2021skeleton}, we randomly select a portion of the labeled data to fine-tune the pre-trained encoder. The average of five runs is reported as the final performance.}\label{tab:sota_semi}
    \begin{tabular*}{0.8\linewidth}{lccccc}
        \toprule[1.2pt]
        \multirow{3}{*}{Method} & \multirow{3}{*}{Venue} & \multicolumn{4}{c}{\textbf{NTU-60}}\\
        \cmidrule[0.8pt](lr){3-6}
        & & \multicolumn{2}{c}{\textbf{x-view}} & \multicolumn{2}{c}{\textbf{x-sub}} \\
        \cmidrule[0.8pt](lr){3-4} \cmidrule[0.8pt](lr){5-6}
        & &   (1\%) & (10\%) & (1\%) & (10\%) \\
        \midrule[0.8pt]
        LongT GAN \citep{zheng2018unsupervised}   & AAAI'2018 &-      &-      &35.2   &62.0   \\
        MS$^2$L \citep{lin_ms2l_mm}               & ACM MM'2020 &-      &-      &33.1   &65.1   \\
        ASSL \citep{si2020adversarial}            & ECCV'2020 &-      &69.8   &-      &64.3   \\
        ISC \citep{thoker2021skeleton}            & ACM MM'2021 &38.1   &72.5   &35.7   &65.9   \\
        CrosSCLR-B \citep{li20213d}               & CVPR’2021 &49.8   &77.0   &48.6   &72.4   \\
        3s-Colorization \citep{yang2021skeleton}  & ICCV'2021 &52.5   &78.9   &48.3   &71.7   \\
        3s-AimCLR \citep{guo2022aimclr}           & AAAI'2022 &54.3   &81.6   &54.8   &78.2   \\
        CPM \citep{zhang2022contrastive}          & ECCV'2022 &57.5   &77.1   &56.7   &73.0   \\
        \midrule[0.8pt]
        \textbf{CMD (Ours)}                      & ECCV'2022 &53.0   &80.2   &50.6   &75.4   \\
        \textbf{I$^2$MD (Ours)}                  & &\textbf{66.9}  &\textbf{84.7}  &\textbf{66.9}  &\textbf{81.1}\\
        \bottomrule[1.2pt]
    \end{tabular*}
\end{table*}

\subsubsection{KNN Evaluation Protocol} An alternative way to use the pre-trained encoder for action classification is to directly apply a K-Nearest Neighbor (KNN) classifier to the learned features of the training samples. Following \citep{su2020predict}, we assign each test sample to the most similar class where its nearest neighbor is in (\emph{i.e.} KNN with k=1). As shown in Table~\ref{tab:sota_comp_knn}, we perform experiments on the NTU-60 and NTU-120 benchmarks and compare the results with previous works. For both datasets, our approaches exhibit the best performance. It is worth noting that the improved version I$^2$MD surpasses CrosSCLR-B \citep{li20213d} by more than 9\% in the more challenging cross-subject and cross-setup protocols.

\subsubsection{Fine-tuned Evaluation Protocol} In fine-tuned evaluation protocol, the encoder is first unsupervisedly pre-trained, then a newly attached linear classifier along with the pre-trained encoder are finetuned for 80 epochs with a learning rate of 0.01 (reduced to 0.001 and 0.0001 at epoch 50 and 70, respectively). As shown in Table~\ref{tab:sota_comp_finetune}, compared to the models with random initialization (w/o pre-training), I$^2$MD pre-training brings significant performance improvements on both NTU-60 and NTU-120 datasets, demonstrating the effectiveness of the proposed pre-training framework.

\subsubsection{Semi-supervised Evaluation Protocol} In semi-supervised classification, both labeled and unlabeled data are included during training. Its goal is to train a classifier with better performance than the one trained with only labeled samples. For a fair comparison, we adopt the same strategy as ISC \citep{thoker2021skeleton}. The pre-trained encoder is fine-tuned together with the post-attached linear classifier on a portion of the corresponding training set. We conduct experiments on the NTU-60 dataset. As shown in Table~\ref{tab:sota_semi}, we report the evaluation results when the proportion of supervised data is set to 1\% and 10\%, respectively. Compared to previous methods like ISC \citep{thoker2021skeleton}, CrosSCLR \citep{li20213d}, Colorization \citep{yang2021skeleton}, AimCLR \citep{guo2022aimclr}, and CPM \citep{zhang2022contrastive}, our algorithm exhibits superior performance. For example, with the same baseline, the proposed CMD outperforms ISC by a large margin.
When the proportion of supervised data is small (1\%), our new I$^2$MD shows an absolute performance improvement of more than 9\% over previous works. Note that in Colorization \citep{yang2021skeleton} and AimCLR \citep{guo2022aimclr}, test-time multimodal ensembling is performed. We can find that our I$^2$MD still outperforms all of these methods without ensembling multiple skeleton modalities.

\begin{table}[t!]
    \centering
    \setlength\tabcolsep{7pt}
    \caption{Performance comparison on PKU-II x-sub in terms of the transfer learning evaluation protocol. The source datasets are NTU-60 x-sub and NTU-120 x-sub.}\label{tab:sota_comp_transfer}
    \begin{tabular*}{1.0\linewidth}{lcc}
        \toprule[1.2pt]
        \multirow{2}{*}{Method}   		  & \multicolumn{2}{c}{\textbf{To PKU-II}} \\
        \cmidrule[0.8pt](lr){2-3}
         & NTU-60          & NTU-120 \\
        \midrule[0.8pt]
        LongT GAN \citep{zheng2018unsupervised}	 & 44.8           & -\\
        MS$^2$L \citep{lin_ms2l_mm}     		     & 45.8           & -\\
        ISC \citep{thoker2021skeleton}    		 & 51.1           & 52.3\\
        CrosSCLR-B		                         & 54.0           & 52.8 \\
        3s-ActCLR \citep{Lin_2023_CVPR}              & 55.9           & -  \\
        \midrule[0.8pt]
        \textbf{CMD (Ours)}               & 56.0  & 57.0\\
        \textbf{I$^2$MD (Ours)}               & \textbf{60.7}  & \textbf{61.8}\\
        \bottomrule[1.2pt]
    \end{tabular*}
\end{table}

\subsubsection{Transfer Learning Evaluation Protocol} In transfer learning evaluation protocol, we examine the transferability of the learned representation. Specifically, we first utilize the proposed framework to pre-train the encoder on the source dataset. Then the pre-trained encoder along with a linear classifier are finetuned on the target dataset for 80 epochs with a learning rate of 0.01 (reduced to 0.001 at epoch 50). We select NTU-60 and NTU-120 as source datasets, and PKU-II as the target dataset. We compare the proposed approach with previous methods like LongT GAN \citep{zheng2018unsupervised}, MS$^2$L \citep{lin_ms2l_mm}, ISC \citep{thoker2021skeleton}, CrosSCLR-B \citep{li20213d}, and ActCLR \citep{Lin_2023_CVPR} under the cross-subject protocol. As shown in Table~\ref{tab:sota_comp_transfer}, both CMD and I$^2$MD exhibit remarkable performance on the PKU-II dataset after large-scale pre-training, outperforming previous methods by a considerable margin. This indicates that the representation learned by our approach is more transferable.

\subsection{Experimental Analysis}
In this section, we conduct several ablative experiments on the NTU-60/NTU-120 datasets. The performance is reported according to the cross-subject protocol. The baseline of our approach is the MoCo v2 \citep{chen2020improved} framework.

\subsubsection{Component Analysis of I $^2$MD}
We first evaluate I$^2$MD and its variants on the NTU-60 and NTU-120 datasets to investigate the effectiveness of each component. As shown in Table~\ref{tab:component_i2md}, the performance under both linear and KNN evaluation protocols is reported. We can find that: \textbf{i)} Either of the two main components in the proposed I$^2$MD leads to considerable performance gains compared to the baseline. Under the linear evaluation protocol, both CMD and IMD bring more than 4\% performance improvement on both NTU-60 and NTU-120 datasets. The performance improvement under the KNN evaluation protocol is even more significant, both exceeding 8\%. \textbf{ii)} Compared to the complete I$^2$MD framework, the removal of either CMD or IMD strategies results in varying degrees of performance degradation, especially under the KNN evaluation protocol. As shown in Table~\ref{tab:component_i2md_transfer}, for each variant, we also report the transferability of the learned representation from the NTU-60 x-sub and NTU-120 x-sub datasets to the smaller PKU-II x-sub dataset, where similar experimental results are also observed.

\begin{figure}[t!]
    \centering
    \includegraphics[width=0.85\linewidth]{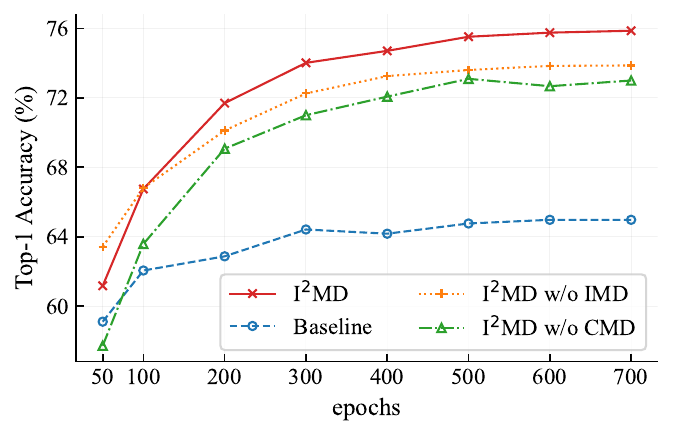}
    \caption{Pre-training schedules for I$^2$MD and its baseline. The results on NTU-60 x-sub according to the KNN evaluation protocol are reported.}\label{fig:convergence}
\end{figure}

In Fig.~\ref{fig:convergence}, we study the pre-training schedules of I$^2$MD variants and the baseline, where KNN-based top-1 accuracy is monitored to observe their respective performance curves. As we can see, a longer pre-training schedule brings more noticeable performance improvement for I$^2$MD and its variants compared to the baseline. After 100 epochs, I$^2$MD consistently outperforms its incomplete variants.

\begin{table}[t!]
    \centering
    \setlength\tabcolsep{3.5pt}
    \caption{Component Analysis of I$^2$MD on NTU-60 x-sub and NTU-120 x-sub. The performance under both linear and KNN evaluation protocols is reported.}\label{tab:component_i2md}
    \begin{tabular*}{1.0\linewidth}{lcccc}
        \toprule[1.2pt]
        \multirow{2}{*}{Method} & \multicolumn{2}{c}{\textbf{Linear}}	& \multicolumn{2}{c}{\textbf{KNN}}\\
        \cmidrule[0.8pt](lr){2-3} \cmidrule[0.8pt](lr){4-5}
        & NTU-60  & NTU-120  & NTU-60  & NTU-120\\
        \midrule[0.8pt]
        Baseline  & 77.8  & 66.7  & 65.0  & 50.8\\
        I$^2$MD w/o CMD & 81.9  & 72.6  & 73.1	& 61.2\\
        I$^2$MD w/o IMD & 83.0  & 71.8  & 73.9	& 59.2\\
        \rowcolor{Gray1}
        I$^2$MD         & \textbf{83.4}  & \textbf{73.1}  & \textbf{75.9}	& \textbf{62.0}\\
        \bottomrule[1.2pt]
    \end{tabular*}
\end{table}

\begin{table}[t!]
    \centering
    \setlength\tabcolsep{9.2pt}
    \caption{Component Analysis of I$^2$MD on PKU-II x-sub according to the transfer learning evaluation protocol. The source datasets are NTU-60 x-sub and NTU-120 x-sub, respectively.}\label{tab:component_i2md_transfer}
    \begin{tabular*}{1.0\linewidth}{lcc}
        \toprule[1.2pt]
        \multirow{2}{*}{Method}   		  & \multicolumn{2}{c}{\textbf{To PKU-II x-sub}} \\
        \cmidrule[0.8pt](lr){2-3}
         & NTU-60 x-sub          & NTU-120 x-sub \\
        \midrule[0.8pt]
        Baseline & 55.1          & 54.3\\
        I$^2$MD w/o CMD     & 59.5          & 59.6\\
        I$^2$MD w/o IMD     & 59.1          & 60.2\\
        \rowcolor{Gray1}
        I$^2$MD             & \textbf{60.7} & \textbf{61.8}\\
        \bottomrule[1.2pt]
    \end{tabular*}
\end{table}

\begin{table*}[t!]
    \centering
    \setlength\tabcolsep{7pt}
    \caption{Ablative experiments of modality selection and bidirectional distillation on NTU-60 x-sub. J, M, and B denote joint, motion, and bone modality, respectively. The horizontal arrows indicate the direction of knowledge distillation. The up arrows indicates the performance improvement compared to the baseline.}
    \label{tab:abla_modality}
    \begin{tabular*}{1.0\linewidth}{lllllll}
        \toprule[1.2pt]
        \multirow{2}{*}{Modality \& Direction} & \multicolumn{3}{c}{\textbf{Linear Evaluation}} & \multicolumn{3}{c}{\textbf{KNN Evaluation}}\\
        \cmidrule[0.8pt](lr){2-4} \cmidrule[0.8pt](lr){5-7}
        & Bone ($\Delta$) & Motion ($\Delta$) & Joint ($\Delta$) & Bone ($\Delta$) & Motion ($\Delta$) & Joint ($\Delta$) \\
        \midrule[0.8pt]
        Baseline 				& 74.4 & 73.1 & 76.1 & 62.0 & 56.8 & 63.4 \\
        \midrule[0.4pt]
        J $\leftarrow$ B     	& 74.4 & -    & 76.5 ($\uparrow$ 0.4) & 62.0 & -   & 64.3 ($\uparrow$ 0.9) \\
        J $\rightleftarrows$ B 	& 76.6 ($\uparrow$ 2.2) & -    & 77.7 ($\uparrow$ 1.6) & 65.9 ($\uparrow$ 3.9) & - & 66.5 ($\uparrow$ 3.1) \\				
        \midrule[0.4pt]
        J $\leftarrow$ M     	& -    & 73.1 &  78.9 ($\uparrow$ 2.8) & - & 56.8 & 64.8 ($\uparrow$ 1.4) \\
        J $\rightleftarrows$ M  & -    & \textbf{77.5} (\textbf{$\uparrow$ 4.4}) & \textbf{79.8} (\textbf{$\uparrow$ 3.7}) & - & 67.0 ($\uparrow$ 10.2) & 68.7 ($\uparrow$ 5.3) \\
        \midrule[0.4pt]
        J $\leftarrow$ M,
        J $\leftarrow$ B     	& 74.4 & 73.1 &  78.8 ($\uparrow$ 2.7) & 62.0 & 56.8  & 66.5 ($\uparrow$ 3.1) \\
        J $\rightleftarrows$ M,
        J $\rightleftarrows$ B,
        M $\rightleftarrows$ B 	& \textbf{77.8} (\textbf{$\uparrow$ 3.4}) & 77.1 ($\uparrow$ 4.0) & 79.4 ($\uparrow$ 3.3) & \textbf{69.5} (\textbf{$\uparrow$ 7.5}) & \textbf{68.7} (\textbf{$\uparrow$ 11.9}) & \textbf{70.6} (\textbf{$\uparrow$ 7.2}) \\
        \bottomrule[1.2pt]
    \end{tabular*}
\end{table*}

\subsubsection{Ablation Study on CMD Part}

To justify the effectiveness of the proposed Cross-modal Mutual Distillation (CMD) strategy, we conduct several ablative experiments on the NTU-60 dataset according to the cross-subject protocol. Since we focus on the CMD part, in this section, the performance of adopting BiGRU as the backbone network is reported, which is consistent with the original conference version \citep{mao2022cmd}.

\begin{figure}[t!]
    \centering
    \includegraphics[width=0.85\linewidth]{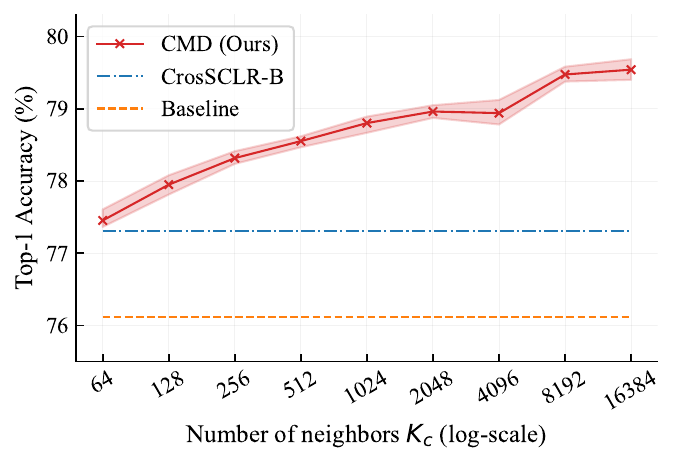}
    \caption{Ablative experiments of the number of neighbors $K_c$ in the cross-modal mutual distillation (CMD) module. The results on NTU-60 x-sub according to the linear evaluation protocol are reported. The solid line and shaded area indicate the mean value and interval of the performance observed in five repeated runs, respectively.}\label{fig:topk}
\end{figure}

\parsection{Impact of Neighbor Number $K_c$}
The number of nearest neighbors controls the abundance of contextual information used in the proposed cross-modal mutual distillation module. We test the performance of the learned representation with respect to different numbers of nearest neighbors $K_c$ under the linear evaluation protocol. As shown in Fig.~\ref{fig:topk}, on the downstream classification task, the performance of the pre-trained encoder improves as $K_c$ increases. When $K_c$ is large enough ($K_c \geq 8192$), continuing to increase its value hardly contributes to the performance. This is because the newly added neighbors are far away and contain little reference value for describing the distribution around the query sample. In addition, we can also find that when the value of $K_c$ varies from 64 to 16384, the performance of our approach is consistently higher than that of CrosSCLR-B \citep{li20213d} and our baseline. This demonstrates the superiority and robustness of the proposed CMD strategy.

\parsection{Modality Selection} In our approach, we consider three kinds of skeleton modalities for self-supervised pre-training as in \citep{li20213d}. They are joint, motion, and bone, respectively. The proposed CMD is capable of performing knowledge distillation between any two of the above modalities. As shown in Table~\ref{tab:abla_modality}, we report the performance of the representation obtained by pre-training with different combinations of skeleton modalities. Note that the joint modality is always preserved since it is used for evaluation. There are several observations as follows:
\textbf{i)} Cross-modal knowledge distillation helps to improve the performance of the representation learned in student modalities.
\textbf{ii)} Under the linear evaluation protocol, mutual knowledge distillation between joint and motion achieves the optimal performance, exceeding the baseline by 3.7\%.
\textbf{iii)} Under the KNN evaluation protocol, the learned representation shows the best results when all the three modalities are involved in mutual knowledge distillation, outperforming the baseline with an absolute improvement of 7.2\%.

\parsection{Bidirectional Distillation} In addition to modality selection, we also verify the effectiveness of bidirectional distillation. It enables the modalities involved in the distillation to interact with each other and progress together, forming a virtuous circle. The results in Table~\ref{tab:abla_modality} show that regardless of which skeleton modality is used during pre-training, bidirectional mutual distillation further boosts the performance, especially under the KNN evaluation protocol.

\parsection{Temperatures in CMD} Table~\ref{tab:abla_temperature} shows the performance of the learned representation for different values of temperature $\tau_t$ and $\tau_s$. We can find that: \textbf{i)} CMD exhibits the optimal performance when using temperature $\tau_s = 0.1$ for the student and a smaller $\tau_t = 0.05$ for the teacher. \textbf{ii)} The performance does not show significant changes when $\tau_t$ varies between small values (from 0.01 to 0.05). \textbf{iii)} The learned representation gets worse as $\tau_s$ increases from 0.1 to 1.0.

\begin{table*}[t!]
    \centering
    \setlength\tabcolsep{6pt}
    \caption{Ablative experiments of the temperatures for teacher and student in CMD. The performance on NTU-60 x-sub according to the linear evaluation protocol is reported.}\label{tab:abla_temperature}
    \begin{tabular*}{0.82\linewidth}{lcccccccccccc}
        \toprule[1.2pt]
        $\tau_t$ & 0.05 & 0.07 & 0.10 & \multicolumn{3}{c}{0.01} & \multicolumn{3}{c}{0.02} & \multicolumn{3}{c}{0.05} \\
        \cmidrule[0.8pt](lr){5-7} \cmidrule[0.8pt](lr){8-10} \cmidrule[0.8pt](lr){11-13}
        $\tau_s$ & 0.05 & 0.07 & 0.10 & 0.1 & 0.5 & 1.0 & 0.1 & 0.5 & 1.0 & 0.1 & 0.5 & 1.0 \\
        \midrule[0.8pt]
        Accuracy (\%) & 77.8 & 78.0 & 77.5 & 79.7 & 78.2 & 76.8 & 79.7 & 77.8 & 76.5 & \textbf{79.8} & 78.0 & 76.8 \\
        \bottomrule[1.2pt]
    \end{tabular*}
\end{table*}

\begin{table}[t!]
    \centering
    \setlength\tabcolsep{6.7pt}
    \caption{One-stage training vs. two-stage training. The first stage of the two-stage training is the standard MoCo v2 \citep{chen2020improved} based contrastive learning. The performance on NTU-60 x-sub according to the linear evaluation protocol is reported.}\label{table:stg}
    \begin{tabular*}{1.0\linewidth}{lcc}
        \toprule[1.2pt]
                    & One-stage	    & Two-stage\\
        \midrule[0.8pt]
        CrosSCLR-B \citep{li20213d}  & 57.6          & 77.3\\
        \rowcolor{Gray1}
        CMD (Ours)  & \textbf{79.4} &\textbf{79.2}\\
        \bottomrule[1.2pt]
    \end{tabular*}
\end{table}

\begin{table}[t!]
    \centering
    \setlength\tabcolsep{13.5pt}
    \caption{Mutual contrastive loss (MCL) vs. mutual distillation (MD)  for the CMD part. The performance on NTU-60 x-sub under linear and KNN evaluation protocols is reported.}\label{table:mcl}
    \begin{tabular*}{1.0\linewidth}{lcc}
        \toprule[1.2pt]
                & Linear & KNN\\
        \midrule[0.8pt]
        Baseline \citep{chen2020improved}   & 76.1              & 63.4\\
        MCL         & 77.3              & 64.3\\
        \rowcolor{Gray1}
        MD         & \textbf{79.4}     & \textbf{70.6}\\
        \bottomrule[1.2pt]
    \end{tabular*}
\end{table}

\parsection{One-stage Training vs. Two-stage Training} We tried one-stage and two-stage training for both CrosSCLR \citep{li20213d} and our CMD. As shown in Table~\ref{table:stg}, CrosSCLR suffers a severe performance drop with one-stage training, while CMD does not show a significant performance gap between one-stage and two-stage training. This demonstrates that the cross-modal positive samples mining strategy in CrossSCLR heavily relies on the initial representation of the model, while our CMD does not show this requirement, reflecting the robustness of CMD to a certain extent.

\parsection{Mutual Contrastive Loss vs. Mutual Distillation} We conduct mutual contrastive learning (MCL in short) for all three modalities and compare its performance with our CMD. As shown in Table~\ref{table:mcl}, both MCL and CMD outperform the baseline, suggesting that the cross-modal interaction is beneficial. Moreover, CMD outperforms MCL under both linear and KNN evaluation protocols, showing the superiority of the proposed mutual knowledge distillation.

\subsubsection{Ablation Study on IMD Part}
In this section, we conduct ablation experiments for the Intra-modal Mutual Distillation (IMD) strategy under the single joint modality. The performance on the NTU-60 dataset according to the cross-subject protocol is reported.

\begin{table}[t!]
    \centering
    \setlength\tabcolsep{7pt}
    \caption{Ablative experiments of the number of neighbors used in the dynamic neighbors aggregation operation. The performance on NTU-60 x-sub under the KNN evaluation protocol is reported.}
    \begin{tabular*}{1.0\linewidth}{lcccc}
        \toprule[1.2pt]
        Number of Neighbors $K_d$  & 16   & 32   & 64 & 128 \\
        \midrule[0.8pt]
        Accuracy (\%)              & 72.8 & 73.0 & \textbf{73.1} & 73.1 \\
        \bottomrule[1.2pt]
    \end{tabular*}
    \label{tab:DNA_number}
\end{table}

\begin{table}[t!]
    \centering
    \setlength\tabcolsep{13.5pt}
    \caption{Mutual contrastive loss (MCL) vs. mutual distillation (MD) for the IMD part. The performance on NTU-60 x-sub under linear and KNN evaluation protocols is reported.}\label{table:cbcl}
    \begin{tabular*}{1.0\linewidth}{lcc}
        \toprule[1.2pt]
                    & Linear	& KNN\\
        \midrule[0.8pt]
        Baseline \citep{chen2020improved}   & 77.8          & 64.5\\
        DNA + MCL  & 79.2          & 68.0\\
        \rowcolor{Gray1}
        DNA + MD   & \textbf{81.9} & \textbf{73.1}\\
        \bottomrule[1.2pt]
    \end{tabular*}
\end{table}

\parsection{Impact of Neighbor Number $K_d$}
For the Dynamic Neighbors Aggregation (DNA) mechanism in IMD, we conducted an experimental analysis of the number of neighbors $K_d$ participating in the local feature aggregation operation. As shown in Table~\ref{tab:DNA_number}, benefiting from the adaptive properties brought by the cross-attention module, the proposed DNA strategy exhibits stable performance for different numbers of neighbors ranging from 16 to 128. When the number of neighbors $K_d$ is set to 64, IMD exhibits the best performance of 73.1\% under the KNN evaluation protocol.

\parsection{Mutual Contrastive Loss vs. Mutual Distillation}
Similar to CMD, in order to transfer knowledge between two branches of the same skeleton modality, an intuitive solution is to construct mutual contrastive learning (MCL in short) across the two branches within each skeleton modality. Therefore, to demonstrate the superiority of cross-branch mutual distillation (MD in short), we also conduct corresponding ablative experiments. As shown in Table~\ref{table:mcl}, DNA with MD significantly outperforms its MCL variant, with absolute performance improvements of 4.1\% and 8.6\% over baseline under linear and KNN evaluation protocols, respectively. This demonstrates the superiority of the proposed mutual knowledge distillation mechanism.

\begin{figure}[t!]
    \centering
    \includegraphics[width=1.0\linewidth]{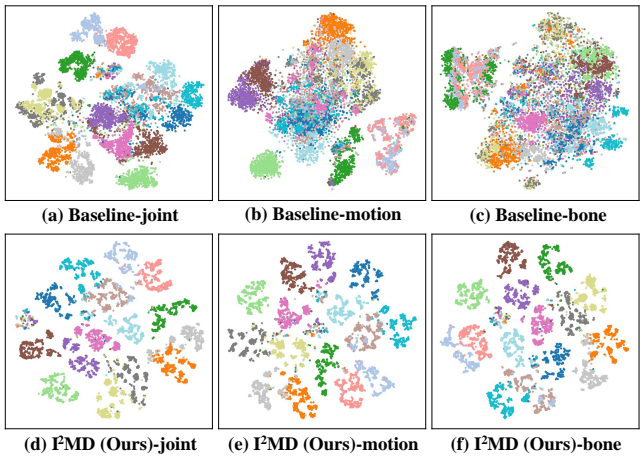}
    \caption{t-SNE \citep{van2008visualizing} visualization of embedding features. We sample 15 action classes from the NTU-60 dataset and visualize the features extracted by the proposed I$^2$MD and its baseline, respectively.}\label{fig:tsne}
\end{figure}

\subsubsection{Qualitative Results}

We visualize the learned representation of the proposed approach and compare it with that of the baseline. The t-SNE \citep{van2008visualizing} algorithm is adopted to reduce the dimensionality of the representation.  To obtain clearer results, we select only 1/4 of the categories in the NTU-60 dataset for visualization. The final results are illustrated in Fig.~\ref{fig:tsne}. For all skeleton modalities, the representation learned by our approach is more compactly clustered than those learned by the baseline in the feature space. This brings a stronger discrimination capability to the representation, explaining the stunning performance of our approach in Table~\ref{tab:sota_comp_knn}.

\section{Conclusion}
In this paper, we present the Inter- and Intra-modal Mutual Distillation (I$^2$MD) framework, aiming to alleviate the limitations of conventional contrastive learning when applied to self-supervised 3D action representation learning. For cross-modal reinforcement, the proposed Cross-modal Mutual Distillation (CMD) reformulates it as a bidirectional knowledge distillation process, where the pairwise similarities between embedding features are utilized to model the modality-specific knowledge. For intra-modal reinforcement, the Dynamic Neighbors Aggregation (DNA) mechanism is first introduced for local cluster-level contrastive learning in parallel with original instance-level discrimination. Intra-modal Mutual Distillation (IMD) is then applied in each skeleton modality for cross-level knowledge exchange. Extensive experiments on three 3D action recognition benchmarks demonstrate the superiority of the proposed I$^2$MD framework.

\section{Data Availability}
The NTU-RGB+D dataset \citep{shahroudy2016ntu,liu2020ntu} and the PKU-MMD \citep{liu2017pku} dataset used in this study are well-recognized public benchmarks in skeleton-based action recognition. The code for data processing has been made publicly available in \url{https://github.com/maoyunyao/CMD}.

\bibliographystyle{spbasic}

\bibliography{reference}

\end{document}